\documentclass[sigconf]{acmart}

\copyrightyear{2023} 
\acmYear{2023}  
\setcopyright{acmlicensed}\acmConference[SIGIR '23]{Proceedings of the 46th
International ACM SIGIR Conference on Research and Development in Information
Retrieval}{July      23--27, 2023}{Taipei, Taiwan}
\acmBooktitle{Proceedings of the 46th International ACM SIGIR Conference on Research
and Development in Information Retrieval (SIGIR '23), July      23--27, 2023,
Taipei, Taiwan}
\acmPrice{15.00}
\acmDOI{10.1145/3539618.3591652}
\acmISBN{978-1-4503-9408-6/23/07}

\usepackage[doipre={doi:~}]{uri}
\AtBeginDocument{%
  \providecommand\BibTeX{{%
    \normalfont B\kern-0.5em{\scshape i\kern-0.25em b}\kern-0.8em\TeX}}}

\usepackage{bm}
\usepackage{color}
\usepackage{enumitem}

\usepackage{hyperref}
\usepackage{tabularx}
\usepackage{subfigure}
\usepackage{multicol}
\usepackage{multirow}
\usepackage{balance}
\usepackage{algorithm}
\usepackage{algorithmicx}
\usepackage[T1]{fontenc}
\usepackage{aecompl}
\usepackage{algpseudocode}

\begin{document}

\title{Continual Learning on Dynamic Graphs via Parameter Isolation}

\author{Peiyan Zhang}
\authornote{Both authors contributed equally to this research.}
\affiliation{\institution{Hong Kong University of \\ Science and Technology}\country{Hong Kong}}
\email{pzhangao@cse.ust.hk}

\author{Yuchen Yan}
\authornotemark[1]
\affiliation{\institution{School of Intelligence Science and Technology, Peking University}\city{Beijing}\country{China}}
\email{2001213110@stu.pku.edu.cn}

\author{Chaozhuo Li}
\authornote{Chaozhuo Li is the corresponding author}
\affiliation{\institution{Microsoft Research Asia}\city{Beijing}\country{China}}
\email{cli@microsoft.com}

\author{Senzhang Wang}
\affiliation{\institution{Central South University}\country{China}}
\email{szwang@csu.edu.cn}

\author{Xing Xie}
\affiliation{\institution{Microsoft Research Asia}\city{Beijing}\country{China}}
\email{xing.xie@microsoft.com}

\author{Guojie Song}
\affiliation{\institution{School of Intelligence Science and Technology, Peking University}\city{Beijing}\country{China}}
\email{gjsong@pku.edu.cn}

\author{Sunghun Kim}
\affiliation{\institution{Hong Kong University of \\ Science and Technology}\country{Hong Kong}}
\email{hunkim@cse.ust.hk}

\renewcommand{\shortauthors}{Peiyan Zhang, et al.}

\begin{abstract}
Many real-world graph learning tasks require handling dynamic graphs where new nodes and edges emerge. Dynamic graph learning methods commonly suffer from the catastrophic forgetting problem, where knowledge learned for previous graphs is overwritten by updates for new graphs. To alleviate the problem, continual graph learning methods are proposed. However, existing continual graph learning methods aim to learn new patterns and maintain old ones with the same set of parameters of fixed size, and thus face a fundamental tradeoff between both goals. In this paper, we propose \textbf{P}arameter \textbf{I}solation \textbf{GNN} (\textbf{PI-GNN}) for continual learning on dynamic graphs that circumvents the tradeoff via parameter isolation and expansion. Our motivation lies in that different parameters contribute to learning different graph patterns. Based on the idea, we expand model parameters to continually learn emerging graph patterns. Meanwhile, to effectively preserve knowledge for unaffected patterns, we find parameters that correspond to them via optimization and freeze them to prevent them from being rewritten. Experiments on eight real-world datasets corroborate the effectiveness of PI-GNN compared to state-of-the-art baselines.
\end{abstract}







\maketitle

\section{Introduction}
\par Graphs are ubiquitous around our life describing a wide range of relational data. Substantial endeavors have been made to learn desirable graph representations to facilitate graph data analysis~\cite{grover2016node2vec,perozzi2014deepwalk,tang2015line,kipf2016semi,hamilton2017inductive,jin2022code,jin2022towards,velivckovic2017graph,huang2022going,yang2022reinforcement,guo2022learning,zhang2023efficiently,guo2022evolutionary}. Graph Neural Networks (GNNs)~\cite{zhao2022learning,zhang2022geometric,pang2022improving,li2022house,ma2022meta,xu2018powerful,wu2019simplifying,guo2023homophily,klicpera2020directional,he2020lightgcn,klicpera2020fast} take advantage of deep learning to fuse node features and topological structures, achieving state-of-the-art performance in a myriad of applications.

\par Conventional GNNs are specifically designed for static graphs, while real-life graphs are inherently dynamic. 
For example, edges are added or removed in social networks when users decide to follow or unfollow others. 
Various graph learning approaches are proposed to tackle the challenge of mutable topology by encoding new patterns into the parameter set, \textit{e.g.,} incorporating the time dimension. Nevertheless, these methods primarily focus on capturing new emerging topological patterns, which might overwhelm the previously learned knowledge, leading to a phenomenon called the \textit{catastrophic forgetting}~\cite{mccloskey1989catastrophic,ratcliff1990connectionist,french1999catastrophic,jin2022prototypical,shi2021overcoming,zhang2023survey}. 
To address catastrophic forgetting in dynamic graphs, \textit{continual graph learning} methods are proposed~\cite{wang2020streaming,zhou2021overcoming} to learn emerging graph patterns while maintaining old ones in previous snapshots. 

Existing continual graph learning methods generally apply techniques such as data replay~\cite{shin2017continual,lopez2017gradient,rebuffi2017icarl,rolnick2018experience,yoon2017lifelong} and regularizations~\cite{kirkpatrick2017overcoming,zenke2017continual,li2017learning,aljundi2018memory,ramesh2021model,ke2020continual,hung2019compacting} to alleviate catastrophic forgetting in dynamic graphs. 
However, these approaches are still plagued by seeking a desirable equilibrium between capturing new patterns and maintaining old ones due to two reasons. 
\begin{itemize}[leftmargin=*,noitemsep,topsep=0pt]
    \item On one hand, existing methods use \textbf{the same parameter set} to preserve both old and new patterns in dynamic graphs. Consequently, new patterns are learned at the inevitable expense of overwriting the parameters encoding old patterns, leading to the potential forgetting of previous knowledge. 
    \item On the other hand, existing methods use parameters with \textbf{fixed size} throughout the dynamic graph. As parameters with a fixed size may only capture a finite number of graph patterns, compromises have to be made between learning new graph patterns and maintaining old ones.
\end{itemize} 

\begin{figure}[t]
\centering
\includegraphics[width=0.40\textwidth]{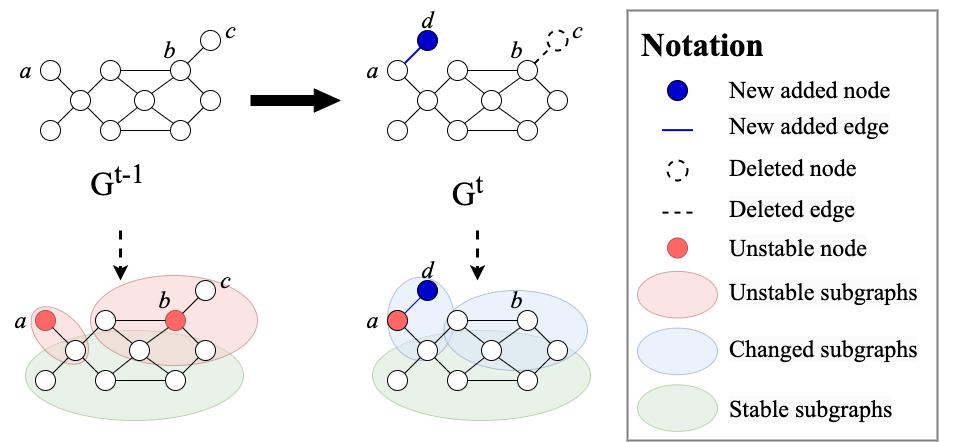} 
\caption{Illustration of stable, unstable, and changed subgraphs with $k=1$. At snapshot $t$, a new node $d$ connects to $a$ and node $c$ is deleted. Thus, the neighborhoods of $a,b$ are unstable subgraphs at snapshot $t-1$ (red shades). Unstable subgraphs will become changed ones after adding and deleting nodes and edges at snapshot $t$ (blue shades). The green shades denote stable subgraphs.}
\label{figure:illustration}
\end{figure}

Moreover, such shared and fixed parameter set might be potentially not suitable in dynamic graph settings. On the one hand, graph evolution can be quite flexible. As shown in Figure~\ref{figure:illustration}, nodes anywhere can be added or deleted, resulting in dramatic changes in each snapshot. On the other hand, unlike non-graph data, \textit{e.g.,} images, where no coupling relationship exists between the data, nodes in the graph can be deeply coupled. As shown in Figure~\ref{figure:illustration}, newly added/dropped nodes will affect the states of their neighborhood nodes, leading to the cascaded changes of node states. 
Therefore, how to perform continual learning in such dramatically and cascadedly changing situation is the unique challenge for continual graph learning. As the emerging patterns in dynamic graphs can be informative, the shared and fixed parameter set will be affected significantly that corrupts old patterns, which inspires us to utilize extra parameters to capture these informative patterns.


Parameter isolation, a continual learning method 
with the intuition that different types of patterns in terms of temporal are captured by different sets of parameters~\cite{rusu2016progressive}, 
 naturally fits the unique challenge of continual graph learning. 
 Thus, to break the tradeoff between learning new patterns and maintaining old ones in dynamic graphs, we propose to utilize parameter isolation to apply separate and expanding parameters to capture the emerging patterns without sacrificing old patterns of the stable graph.



In particular, applying parameter isolation in continual graph learning is challenging. First of all, existing parameter isolation approaches~\cite{rusu2016progressive,rosenfeld2018incremental,yoon2017lifelong,fernando2017pathnet} are for data in Euclidean space, \textit{e.g.,} images, which cannot capture the influence of cascading changes in dynamic graphs well. Second, during the graph evolution process, parameter isolation methods continually expand the model scale and suffer the dramatically increased parameters~\cite{rusu2016progressive}. Last but not least, as existing methods cannot provide theoretical guarantees to ensure the ability of the parameter isolation process, unstable and inconsistent training may occur due to the dynamic data properties.

To address these issues, we propose a novel  \textbf{P}arameter \textbf{I}solation \textbf{GNN} (\textbf{PI-GNN}) model for continual learning on dynamic graphs. To adapt to the unique challenges of continual graph learning, at the data level, the whole graph pattern is divided into stable and unstable parts to model the cascading influences among nodes. At the model level, 
PI-GNN is designed to consist of two consecutive stages: the \textit{knowledge rectification} stage and the \textit{parameter isolation} stage. As the foundation for parameter isolation, the knowledge rectification stage filters out the impact of patterns affected by changed nodes and edges. We fine-tune the model to exclude such impact and get stable parameters that maintain performance over existing stable graph patterns. In the parameter isolation stage, we keep the stable parameters frozen, while expanding the model with new parameters, such that new graph patterns are accommodated without compromising stable ones. To prevent the model from excessive expansion, we use model distillation to compress the model while maintaining its performance. Finally, we provide theoretical analysis specifically from the perspective of graph data to justify the ability of our method. We theoretically show that the parameter isolation stage optimizes a tight upper bound of the retraining loss on the entire graph. Extensive experiments are conducted on eight real-world datasets, where PI-GNN outperforms state-of-the-art baselines for continual learning on dynamic graphs. We release our code at \href{https://github.com/Jerry2398/PI-GNN}{https://github.com/Jerry2398/PI-GNN}.

\par Our contributions are summarized as follows:
\begin{itemize}[leftmargin=*,noitemsep,topsep=0pt]
    \item We propose PI-GNN, a continual learning method on dynamic graphs based on parameter isolation. By isolating stable parameters and learning new patterns with parameter expansion, PI-GNN not only captures emerging patterns but also suffers less from catastrophic forgetting. 
    \item We theoretically prove that PI-GNN optimizes a tight upper bound of the retraining loss on the entire graph.
    \item We conduct extensive evaluations on eight real-world datasets, where PI-GNN achieves state-of-the-art on dynamic graphs. 
\end{itemize}

\section{Preliminaries}
In this section, we first present preliminaries on dynamic graphs, and then formulate the problem definition.

\noindent\textbf{Definition 1 (Dynamic Graph).}
A graph at snapshot $t$ is denoted as $G^{t}=(V^t, \mathbf{A}^{t},\mathbf{X}^{t})$, where $V^t$ is the set of nodes, $\mathbf{A}^{t}$ is the adjacency matrix, and $\mathbf{X}^{t}$ are node features. A dynamic graph is denoted as $\mathcal{G}=(G^{1},...,G^{T-1},G^{T})$, with $T$ the number of snapshots. 

\noindent\textbf{Definition 2 (Graph Neural Networks (GNNs)).}
We denote a $k$-layer GNN parameterized by $\theta^{t}$ as $f_{\theta^{t}}$, the representations of node $v$ in layer $l$ as $\mathbf{h}_{v}^{(l)}$ and $\mathbf{h}_{v}^{(0)}=\mathbf{X}_v$. The representations of node $v$ is updated by message passing as follows:
\begin{equation}
\begin{aligned}
    \mathbf{h}_{v}^{(l+1)} = \mathrm{COMB}\left(\mathrm{AGG}\left(\mathbf{h}_{u}^{(l)}, \forall u \in N(v)\right), \mathbf{h}_{v}^{(l)}\right), 
\end{aligned}
\end{equation}
where $N(v)$ is the neighborhood of node $v$, AGG and COMB denote GNNs' message passing and aggregation mechanisms, respectively. 

With $y_v$ as the label of $v$, the loss of node classification is:
\begin{equation}
\begin{aligned}
     L \left (f_{\theta^t}\left(G^t\right) \right ) = \sum_{v \in V^{t}} \mathrm{CrossEnt}\left(\mathrm{softmax} \left(\mathbf{h}_{v}^{(k)} \right), y_{v} \right).
\end{aligned}
\end{equation}

In dynamic graphs, some subgraphs at snapshot $t$ will be affected by new or deleted graph data at $t+1$~\cite{pareja2020evolvegcn}. Based on this motivation, we decompose the graph into stable and unstable subgraphs: 
\begin{itemize}[leftmargin=*,noitemsep,topsep=0pt]
    \item \textbf{Unstable subgraphs $G_{unstable}^{t-1}$} are defined as the $k$-hop ego-subgraphs at snapshot $t-1$ that centered with nodes directly connected to emerging/deleted nodes or edges at snapshot $t$.
    \item \textbf{Stable subgraphs $G_{stable}^{t-1}$} are defined as the $k$-hop ego-subgraphs whose center nodes are outside the $k$-hop ranges of nodes connecting to emerging/deleted nodes or edges at snapshot $t$, i.e. $G_{stable}^{t-1} = G^{t-1}-G_{unstable}^{t-1}$.
    \item \textbf{Changed subgraphs $\Delta G^{t}$} evolve from $G_{unstable}^{t-1}$ after new nodes and edges are added or deleted at snapshot $t$.
\end{itemize}

Note that $G_{stable}$ and $G_{unstable}$ refers to ego-networks while the operation conducted on $G_{stable}$ and $G_{unstable}$ only involves their central nodes. As the node is either stable or unstable, $G_{stable} = G-G_{unstable}$ will hold. Although the ego-networks might be overlapping, this won’t affect the above equation.

Based on these definitions, we have
\begin{equation}
\begin{aligned}
    & L \left (f_{\theta^{t-1}}\left(G^{t-1}\right) \right ) \\
    & = L \left (f_{\theta^{t-1}}\left( G_{stable}^{t-1}\right) \right ) + L \left (f_{\theta^{t-1}}\left(G_{unstable}^{t-1}\right) \right ), \\
\end{aligned}
\label{equ:relation_1}
\end{equation}

Note that the loss of node classification is computed in units of nodes. Therefore, the loss function in r.h.s of Eqn.~\ref{equ:relation_1} can be directly added up so that Eqn.~\ref{equ:relation_1} will hold.

As $G_{stable}^{t-1}$ will remain unchanged at snapshot $t$ while $G_{unstable}^{t-1}$ will evolve into $\Delta G^{t}$, we have
\begin{equation}
\begin{aligned}
    L \left (f_{\theta^t}\left(G^{t}\right) \right ) = L \left (f_{\theta^t}\left(G_{stable}^{t-1}\right) \right ) + L \left (f_{\theta^t}\left(\Delta G^{t}\right) \right )
\end{aligned}
\label{equ:relation_2}
\end{equation}

We finally define the problem of continual learning for GNNs.

\noindent\textbf{Definition 3 (Continual Learning for GNNs).}
The goal of continual learning for GNNs is to learn a series of parameters $\theta=(\theta^{1},\theta^{2},...,\theta^{T})$ given a dynamic graph $\mathcal{G}=(G^{1}, G^{2},..., G^{T})$, where $f_{\theta^t}$ perform well on $G^{i \leq t}$, i.e. graphs from all previous snapshots. 


\begin{figure*}[h]
\centering
\includegraphics[width=0.8\textwidth]{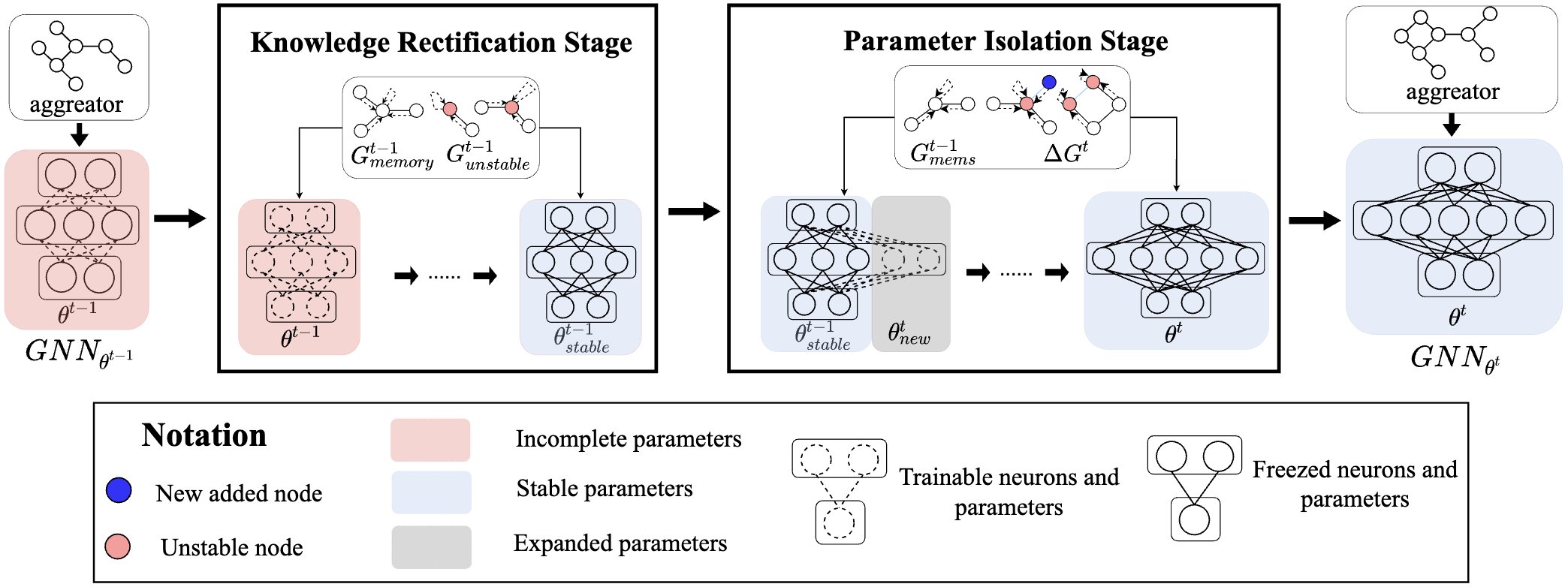}
\caption{An overview of PI-GNN. There are two consecutive stages in PI-GNN: the knowledge rectification stage and the parameter isolation stage. The knowledge rectification stage is used to obtain stable model parameters, while the parameter isolation stage is used to learn new patterns without compromising existing stable patterns.}
\label{figure:framework}
\end{figure*}


\section{Methodology}
In this section, we introduce the details of PI-GNN, as shown in Figure \ref{figure:framework}. PI-GNN contains two stages. 
In the \textit{knowledge rectification} stage, the GNN model is trained to learn stable paramters to capture topological structure within the stable subgraphs.  In the \textit{parameter isolation} stage, the GNN model is further expanded to capture the new patterns within the unstable subgraphs while the stable parameters are fixed from being rewritten. 
Theoretical analysis are presented to justify the capacity of both stages to comprehensively capture new and stable patterns at each snapshot.  

\subsection{Knowledge Rectification}
The knowledge rectification stage is used to update the model to obtain parameters that capture the topological patterns of stable subgraphs. The design philosophy behind the knowledge rectification is that $\theta^{t-1}$ may not be the optimal initialization for snapshot $t$ because they may encode irrelevant data patterns.  Specifically, as nodes are connected with each other, new nodes and edges at snapshot $t$ will have an influence on existing nodes at $t-1$, which constitute unstable subgraphs $G_{unstable}^{t-1}$ that change at snapshot $t$. Thus, only the stable subgraphs $G_{stable}^{t-1}$ should be shared between $G^{t-1}$ and $G^t$, and consequently, we need to initialize $\theta^t$ that best fits $G_{stable}^{t-1}$ instead of $G^{t-1}$. We achieve this goal by the following optimization. From Eqn. \ref{equ:relation_1}, we have:
\begin{equation}
\begin{aligned}
    & \min_{\theta^{t-1}} \; L\left(f_{\theta^{t-1}}\left(G_{stable}^{t-1}\right)\right) \\ 
    \Leftrightarrow & \min_{\theta^{t-1}} \; L\left(f_{\theta^{t-1}}\left(G^{t-1}\right)\right) - L\left(f_{\theta^{t-1}}\left(G_{unstable}^{t-1}\right)\right). \\
\end{aligned}
\label{opt:stable_knowledge}
\end{equation}

By optimizing the model via Eqn. \ref{opt:stable_knowledge}, we get parameters encoding $G_{stable}^{t-1}$ which will remain unchanged at snapshot $t$. We denote them as \textit{stable parameters}. In Eqn. \ref{opt:stable_knowledge}, $G_{unstable}^{t-1}$ can be easily obtained from $\Delta G^{t}$ through its definition. 
However, it is time-consuming to re-compute $L(f_{\theta^{t-1}}(G^{t-1}))$. We thus sample and store some nodes from $G^{t-1}$ and use their $k$-hop neighborhoods (referred to as $G_{memory}^{t-1}$) to estimate $L(f_{\theta^{t-1}}(G^{t-1}))$. We adopt uniform sampling to ensure unbiasedness. Therefore, in practice, the following optimization problem is solved for the knowledge rectification stage,
\begin{equation}
\begin{aligned}
\label{equ:rectify}
\theta^{t-1}_{stable} = \arg \min_{\theta^{t-1}} \; & L(f_{\theta^{t-1}}(G_{memory}^{t-1})) \\
 - & \beta L(f_{\theta^{t-1}}(G_{unstable}^{t-1})),
\end{aligned}
\end{equation}
where $\theta_{stable}^{t-1}$ are the stable parameters, $\beta$ is a factor to handle the size imbalance problem caused by sampling $G_{memory}^{t-1}$.

\subsection{Parameter Isolation}
After obtaining the stable parameters, we propose a parameter isolation algorithm for GNNs to learn new patterns incrementally without compromising the stable patterns.

\subsubsection{Algorithm}

\begin{figure}[t]
\centering
\includegraphics[width=0.36\textwidth]{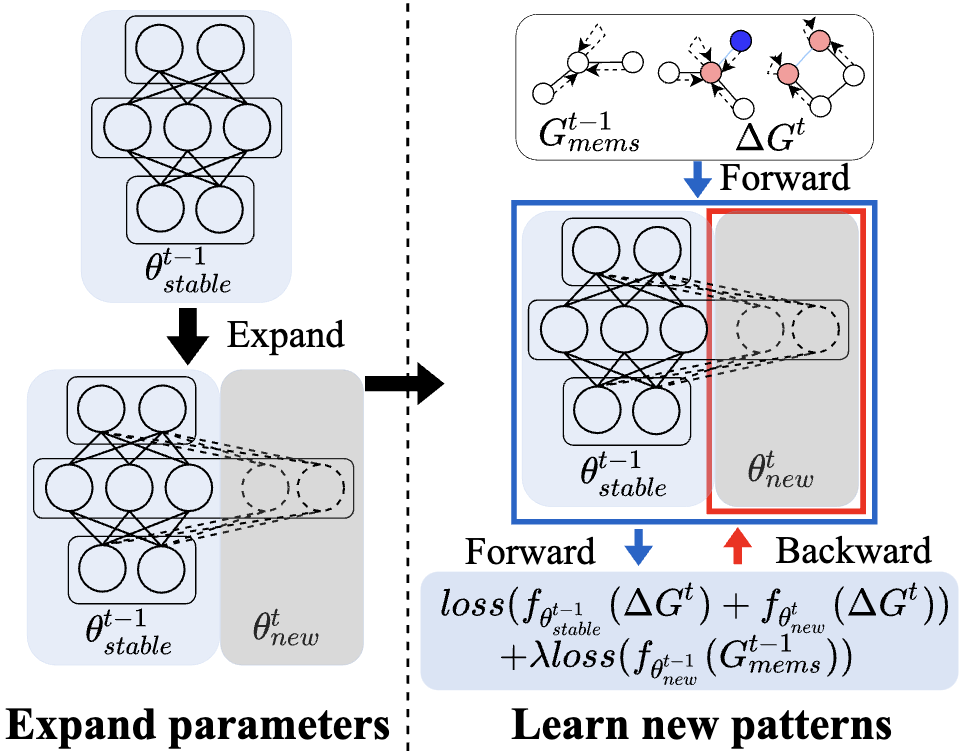} 
\caption{Illustration of parameter isolation stage.}
\label{figure:parameter_isolation_framewwork}
\end{figure}

The key idea of parameter isolation is that different parameters can learn different patterns, and that instead of modifying existing parameters, we isolate and add new parameters to learn new patterns. Specifically, we show the parameter isolation algorithm at snapshot $t$ in Figure \ref{figure:parameter_isolation_framewwork}, which can be divided as follows:

\begin{enumerate}
    \item \textit{Expand}: We expand parameters for hidden layers $\theta_{stable}^{t-1}$ to $\theta^{t} = \theta_{stable}^{t-1} \oplus \theta_{new}^{t}$, so as to increase the model capacity to learn new patterns. 
    \item \textit{Freeze}: We freeze $\theta_{stable}^{t-1}$ to prevent them from being rewritten. Considering that old patterns may benefit learning new graph patterns, $\theta_{stable}^{t-1}$ are used to compute the loss during the forward pass but will not be updated. 
    \item \textit{Update}: We update $\theta_{new}^t$ on the graph $G^t$. For efficient training, $\theta_{new}^{t}$ are updated only on $\Delta G^{t}$ and $G^{t-1}_{memory}$. Formally, we optimize the following objective
\begin{equation}
\begin{aligned}
    \label{parameters isolation optim}
    \min_{\theta_{new}^{t}} \; & L\left(f_{\theta^{t-1}_{stable}}\left( \Delta G^{t}\right) +f_{\theta_{new}^t}\left(\Delta G^t\right)\right)
    \\&+ \lambda \, L(f_{\theta_{new}^{t}}(G_{memory}^{t-1})),
\end{aligned}
\end{equation}
where $G_{memory}^{t-1}$ is the stable subset of $G_{memory}^{t-1}$, and $\lambda$ is similar to $\beta$ in Eqn. \ref{equ:rectify} that handles the size imbalance problem caused by sampling $G_{memory}^{t-1}$.  
\end{enumerate}

\subsection{Theoretical Analysis}
We conduct theoretical analysis to guarantee the correctness of the proposed parameter isolation algorithm on dynamic graphs. We first present the main theorem as follows:

\begin{theorem}
\begin{equation}
\small
\begin{aligned}
      L\left(f_{\theta^t}\left(G^t\right)\right)&\le L\left(f_{\theta^{t-1}_{stable}}\left(\Delta G^{t}\right) +  f_{\theta_{new}^{t}}\left(\Delta G^{t}\right)\right)\\
      & + \frac{1}{2}L\left(f_{\theta_{stable}^{t-1}}\left(G_{stable}^{t-1}\right)\right) + \frac{1}{2}L\left(f_{\theta_{new}^{t}}\left(G_{stable}^{t-1}\right)\right)
      \label{opt:target} 
\end{aligned}
\end{equation}
and equality holds when $f_{\theta_{new}^t}(G_{stable}^{t-1}) = f_{\theta_{stable}^{t-1}}(G_{stable}^{t-1})$. 
\label{thm}
\end{theorem}

The implication of Theorem \ref{thm} is that by splitting $\theta^t=\theta_{new}^{t}\oplus \theta^{t-1}_{stable}$ and $G^t=G^{t-1}_{stable}\cup \Delta G^t$, we actually optimize a tight upper bound (R.H.S. of Eqn. \ref{opt:target}) of the retraining loss $L(f_{\theta^t}(G^t))$ at snapshot $t$. To prove the theorem, we first show a lemma whose proof can be found in the Appendix \ref{app:lemma_proof}. 
\begin{lemma}
$2L(f_{\theta_{1}}(G)+f_{\theta_{2}}(G)) \leq L(f_{\theta_{1}}(G)) + L(f_{\theta_{2}}(G))$ where the equality holds if $f_{\theta_{1}}(G)=f_{\theta_2}(G)$.
\label{lem}
\end{lemma}

Lemma \ref{lem} essentially follows from the convexity of the cross entropy loss. We then provide proof to Theorem \ref{thm}. 
\begin{proof}
From Eqn. \ref{equ:relation_2}, $\theta^{t} = \theta_{stable}^{t-1} \oplus \theta_{new}^{t}$ and Lemma \ref{lem},
\begin{equation}
\begin{aligned}
& \; L\left(f_{\theta^{t}}\left(G^{t}\right)\right) \\
= & \; L\left(f_{\theta^{t}}\left(\Delta G^{t}\right)\right) + L\left(f_{\theta^{t}}\left(G^{t-1}_{stable}\right)\right) \\
= & \; L\left(f_{\theta^{t-1}_{stable}}\left(\Delta G^{t}\right) + f_{\theta_{new}^{t}}\left(\Delta G^{t}\right)\right) \\
& + L\left(f_{\theta^{t-1}_{stable}}\left(G^{t-1}_{stable}\right) + f_{\theta_{new}^{t}}\left(G^{t-1}_{stable}\right)\right) \\
\leq & \; L\left(f_{\theta^{t-1}_{stable}}\left(\Delta G^{t}\right) + f_{\theta_{new}^{t}}\left(\Delta G^{t}\right)\right) \\
& + \frac{1}{2} L\left(f_{\theta^{t-1}_{stable}}\left(G^{t-1}_{stable}\right)\right) + \frac{1}{2} L\left(f_{\theta_{new}^{t}}\left(G^{t-1}_{stable}\right)\right). \\
\end{aligned}
\label{eq:total_unfold}
\end{equation}
From Lemma \ref{lem}, equality holds if $f_{\theta_{new}^t}(G_{stable}^{t-1}) = f_{\theta_{stable}^{t-1}}(G_{stable}^{t-1})$. Therefore we prove Theorem \ref{thm}.
\end{proof}

\subsubsection{Practical Implementation} We discuss how to optimize the upper bound (R.H.S. of Eqn. \ref{opt:target}) in practice. The second term\\ $L(f_{\theta^{t-1}_{stable}}(G^{t-1}_{stable}))$ has been optimized in Eqn. \ref{equ:rectify} and $\theta_{stable}^{t-1}$ are fixed. Therefore, we only need to optimize the first and third terms. 
\par For the first term of the upper bound, as $\theta^{t-1}_{stable}$ are frozen to preserve the stable parameters, we only optimize on $\theta_{new}^t$. Thus, we optimize the first term as
\begin{equation}
\label{equ:incr}
\min_{\theta_{new}^{t}} \; L\left(f_{\theta^{t-1}_{stable}}\left(\Delta G^{t}\right) + f_{\theta_{new}^{t}}\left(\Delta G^{t}\right)\right).
\end{equation}
\par For the third term, we use $G_{memory}^{t-1}$ to estimate $G^{t-1}_{stable}$. Thus, the third term can be approximately optimized as
\begin{equation}
\label{equ:comp}
\min_{\theta_{new}^{t}} \lambda L(f_{\theta_{new}^{t}}(G_{memory}^{t-1})),
\end{equation}
where $\lambda$ is a balance factor to compensate for the size difference between $G_{memory}^{t-1}$ and $G_{stable}^{t-1}$. Combining Eqn. \ref{equ:incr} and Eqn. \ref{equ:comp}, we obtain the final optimization problem in Eqn. \ref{parameters isolation optim}. 

\subsubsection{Bound Analysis} From Lemma \ref{lem}, the equality holds if\\ $f_{\theta_{new}^t}(G_{stable}^{t-1}) = f_{\theta_{stable}^{t-1}}(G_{stable}^{t-1})$. As ${\theta_{stable}^{t-1}}$ has been optimized in the knowledge rectification stage, $\theta_{stable}^{t-1}$ minimizes $L(f_{\theta^{t-1}}(G_{stable}^{t-1}))$. In addition, as we minimize $L(f_{\theta_{new}^{t}}(G_{memory}^{t-1}))$ and $G_{memory}^{t-1}$ are sampled from $G_{stable}^{t-1}$, we also push $\theta_{new}^t$ close to the minimizer $\theta_{stable}^{t-1}$. Thus, the tightness of the bound in Eqn. \ref{parameters isolation optim} can be ensured.

\subsection{Discussions}
\textbf{Model Compression.} Parameter isolation suffers the dramatically increased parameters for large $T$~\cite{rusu2016progressive}. We adopt model distillation \cite{hinton2015distilling} to solve this problem, where we use the original model as the teacher model and another model with fewer hidden units as the student model (i.e., the compressed model). We minimize the cross entropy between the outputs of the teacher and student models. Specifically, we use $G_{memory}$ as the stable data and $\Delta{G}$ as the new data, altogether severs as our training data for distillation. This is in line with our settings, as these data can be obtained in the knowledge rectification stage, and no additional knowledge is introduced.

Another issue is that if graph patterns disappear significantly due to node/edge deletions, an overly expanded model may face the overfitting problem. This can also be solved by model distillation. When too many nodes or edges are deleted in dynamic graphs compared to the previous snapshot, we can distill the model to get a smaller model at this snapshot to avoid the overfitting problem. 

In our experiments, we compress our model on graph $G^{T}$ after the last snapshot.  As there is no end on the snapshots in real-world applications, the model can be compressed regularly at any time and this barely affect the performance. 

\noindent\textbf{Extension to Other Tasks.}
Throughout the paper, we mainly focus on node classification tasks as in \cite{wang2020streaming,zhou2021overcoming}. However, the theory stated in Theorem \ref{thm} holds for other tasks as long as the loss function is convex. Parameter isolation for other tasks (e.g. link prediction) may require minor design modifications (e.g. graph decomposition, choice of samples), and thus we leave them as future work.



\section{Experiments}

\begin{table}[t]
\small
\centering

\begin{tabular}{c|cccc}
\toprule
\multicolumn{1}{l|}{} & Nodes  & Edges  & Attributes & \# Tasks \\
\midrule
Cora                  & 2,708  & 5,429   & 1,433      & 14          \\
Citeseer              & 3,327  & 4,732   & 3,703      & 12          \\
Amazon Photo          & 7,650  & 119,043 & 745        & 16          \\
Amazon Computer          & 13,752  & 491,722 & 767        & 18          \\
Elliptic          & 31,448  & 23,230 & 166        & 9          \\
DBLP-S                  & 20,000 & 75,706 & 128        & 26          \\
Arxiv-S                 & 40,726  & 88,691  & 128        & 13          \\
Paper100M-S             & 49,459  & 108,710 & 128        & 11          \\
\bottomrule
\end{tabular}
\caption{Statistics of the eight datasets.}
\label{table:dataset summary}
\end{table}

\par In this section, we present the empirical evaluations of PI-GNN on eight real-world datasets ranging from citation networks~\cite{mccallum2000automating,sen2008collective,tang2008arnetminer,wang2020microsoft} to bitcoin transaction networks~\cite{weber2019anti} and e-commercial networks~\cite{shchur2018pitfalls}. 

\subsection{Experimental Setup}
\subsubsection{Datasets}
We conduct experiments on eight real-world graph datasets with diverse application scenarios: Cora~\cite{mccallum2000automating}, Citeseer \cite{sen2008collective}, Amazon Photo \cite{shchur2018pitfalls}, Amazon Computer~\cite{shchur2018pitfalls}, Elliptic~\cite{weber2019anti}, DBLP-S \cite{tang2008arnetminer}, Arxiv-S, and Paper100M-S \cite{wang2020microsoft}. Among them, Elliptic, DBLP-S, Arxiv-S, and Paper100M-S are streaming graphs. We sample from their full data and we split them into tasks according to the timestamps. For Arxiv-S, we choose a subgraph with 5 classes whose nodes number have top variations from year 2010 to year 2018 from Arxiv full data. We split it into tasks according to the timestamps. For DBLP-S, we randomly choose 20000 nodes with 9 classes and 75706 edges from DBLP full data, we split it into tasks according to the timestamps. For Paper100M-S, we randomly choose 12 classes from year 2009 to year 2019 from Paper100M full data and we split it into tasks according to the timestamps. Alternatively, as Cora, Citeseer, Amazon Photo and Amazon Computer are static graphs, we manually transform them into streaming graphs according to the task-incremental setting, where we add one class every two tasks.

For each dataset, we split 60\% of nodes in $G^T$ for training, 20\% for validation and 20\% for test. The train/val/test split is consistent across all snapshots. Dataset statistics are shown in Table \ref{table:dataset summary}.


\subsubsection{Baselines}
We compare PI-GNN with the following baselines. 
\begin{itemize}[leftmargin=*,noitemsep,topsep=0pt]
     \item \textit{Retrain}: We retrain a GNN on $G^t$ at each snapshot. As all patterns are learned for each snapshot, it can be seen as an upper bound for all continual learning methods.
     \item \textit{Pretrain} We use a GNN trained on $G^1$ for all snapshots.
     \item \textit{OnlineGNN}: We train a GNN on $G^1$ and fine tune it on the changed subgraphs $\Delta G^{t}$ at each snapshot $t$.
     \item \textit{Dynamic baselines}: We compare PI-GNN with the following methods on general dynamic graph learning: EvolveGCN~\cite{pareja2020evolvegcn}, DyGNN~\cite{ma2020streaming}, and DNE~\cite{du2018dynamic}.
     \item \textit{Continual baselines}: We compare PI-GNN with the following methods on continual graph learning: ContinualGNN ~\cite{wang2020streaming}, DiCGRL ~\cite{kou2020disentangle}, and TWP ~\cite{liu2021overcoming}.

 \end{itemize}

\subsubsection{Hyperparameter Settings}
\par GraphSAGE
with $k=2$ layers is used as the backbone of \textit{Retrain}, \textit{OnlineGNN}, \textit{Pretrain} and \textit{PI-GNN}. We initialize 12 units for each hidden layer and expand 12 units at each snapshot. We set $\lambda=0.1, \beta=0.01$. We sample 128 nodes for Arxiv-S as $G_{memory}$ and 256 nodes for other datasets. After training, we distill PI-GNN to a smaller model with 32 hidden units. 
We apply grid search to find the optimal hyper-parameters for each model. For fairness, \textbf{all the baselines use backbone models with the final (i.e. largest) capacity of PI-GNN before distillation} to eliminate the influence of different parameter size of models. 

\begin{table*}[t]
\huge
\centering
\renewcommand{\arraystretch}{1.5}
\resizebox{\linewidth}{!}{
\begin{tabular}{c|c|c|c|ccc|ccc|c|cc}
\toprule
\multicolumn{2}{c|}{$\ $}  & Lower bound  & Incremental  & \multicolumn{3}{c|}{Dynamic}   & \multicolumn{3}{c|}{Continual}   &   Upper bound   &  \multicolumn{2}{c}{Ours}  \\

\midrule
                           Dataset      & Metric           &  Pretrain        & OnlineGNN  &   EvolveGCN    &    DyGNN   &   DNE 
                           & ContinualGNN & DiCGRL  & TWP & Retrain & PI-GNN & Distilled PI-GNN   \\
\midrule
                           
\multirow{2}{*}{Arxiv-S}     & PM        & 45.99$\pm$1.99\%.  & 72.69$\pm$1.31\%  & 74.08$\pm$0.35\%  & 61.13$\pm$2.97\%  & 39.13$\pm$5.97\%   & 75.36$\pm$0.21\%    &  67.93$\pm$0.97\%   &  73.91$\pm$0.14\%   &  \textbf{81.86$\pm$0.28\%}   & \textbf{77.02$\pm$0.53\%}   &  76.50$\pm$0.38\%    \\
                          & FM        & 0$\pm$0\%.  & -8.57$\pm$1.77\%   & -5.55$\pm$0.33\%  & -6.98$\pm$2.18\%  & -16.98$\pm$2.18\%   & -2.17$\pm$0.07\%   &  -2.91$\pm$0.44\%   &  -2.12$\pm$0.29\%   &  \textbf{-0.12$\pm$0.06\%}   &  \textbf{-1.22$\pm$0.03\%}   &  -1.71$\pm$0.20\%    \\
\midrule
\multirow{2}{*}{DBLP-S}     & PM    &  43.17$\pm$1.45\%   &  63.52$\pm$1.07\%
                            &  61.52$\pm$1.34\%   &  54.32$\pm$2.75\%   &                                32.44$\pm$6.17\%   &  62.67$\pm$0.06\%   &  57.09$\pm$4.13\%   &  55.82$\pm$4.73\%                              &  \textbf{64.77$\pm$1.01\%}   &  \textbf{64.76$\pm$0.19\%}   &  63.49$\pm$0.45\% \\

                            & FM  &  0$\pm$0\%   &  -5.04$\pm$0.46\%  &  -4.91$\pm$0.23\%     &  -8.62$\pm$2.33\%     &  -15.62$\pm$5.01\%     &  \textbf{-0.13$\pm$0.03\%}     &  -3.24$\pm$0.82\%     &  -5.24$\pm$0.73\%     &  \textbf{1.97$\pm$0.13\%}     &  -0.64$\pm$0.07\%     &  -1.11$\pm$0.15\%   \\
\midrule
\multirow{2}{*}{Paper100M-S}     & PM     &  55.13$\pm$2.26\%    &  77.70$\pm$0.28\%    &  78.14$\pm$0.61\%    &  69.33$\pm$6.91\%    &  42.53$\pm$9.05\%    &  80.59$\pm$0.17\%    &  OOM    &  65.48$\pm$0.43\%    &  \textbf{83.42$\pm$0.05\%}    &  \textbf{83.05$\pm$0.26\%}    &  82.82$\pm$0.18\%  
\\
& FM     &  0$\pm$0\%    &  -6.62$\pm$1.23\%    &  -3.17$\pm$0.61\%    &  -5.25$\pm$1.88\%    &  -11.71$\pm$6.56\%    &  -2.02$\pm$0.02\%    &  OOM    &  -4.86$\pm$1.60\%    &  \textbf{-0.64$\pm$0.07\%}    &  \textbf{-0.93$\pm$0.21\%}    &  -1.49$\pm$0.56\%  \\
\midrule
\multirow{2}{*}{Elliptic}     & PM     &  81.31$\pm$3.97\%    &  78.73$\pm$2.14\%    &  86.71$\pm$0.74\%    &  83.46$\pm$2.28\%    &  67.05$\pm$5.16\%    &  92.12$\pm$0.58\%    &  84.39$\pm$0.35\%    &  89.77$\pm$2.11\%    &  \textbf{94.08$\pm$0.98\%}    &  \textbf{93.65$\pm$1.15\%}    &  92.36$\pm$0.50\%  
\\
& FM     &  0$\pm$0\%    &  -6.91$\pm$1.85\%    &  -4.29$\pm$0.36\%    &  -5.72$\pm$1.61\%    &  -8.43$\pm$5.18\%    &  -0.56$\pm$0.13\%    &  -1.74$\pm$0.49\%    &  -2.05$\pm$0.40\%    &  \textbf{1.21$\pm$0.26\%}    &  \textbf{0.25$\pm$0.10\%}    &  -0.66$\pm$0.21\%  \\
\midrule
\multirow{2}{*}{Cora}     & PM    &  53.65$\pm$1.99\%    &  64.82$\pm$1.31\%    &  62.22$\pm$4.67\%    &  55.37$\pm$4.16\%    &  50.50$\pm$6.29\%    &  74.65$\pm$0.35\%    &  62.93$\pm$0.97\%    &  77.05$\pm$0.14\%    &  \textbf{81.04$\pm$0.28\%}    &  77.63$\pm$0.53\%    &  76.48$\pm$0.66\%  
\\
& FM    &  0$\pm$0\%    &  -6.01$\pm$2.77\%    &  -3.14$\pm$1.71\%    &  -7.20$\pm$1.63\%    &  -18.47$\pm$5.31\%    &  -1.51$\pm$0.07\%    &  -0.84$\pm$0.04\%    &  -1.31$\pm$0.29\%    &  -0.51$\pm$0.19\%    &  \textbf{1.66$\pm$0.16\%}    &  \textbf{1.07$\pm$0.09\%}
\\
\midrule
\multirow{2}{*}{Citeseer}   & PM    &  38.49$\pm$0.45\%    &  59.58$\pm$1.56\%    &  67.12$\pm$1.01\%    &  66.28$\pm$3.62\%    &  51.95$\pm$7.25\%    &  68.92$\pm$0.29\%    &  66.60$\pm$0.05\%    &  69.24$\pm$0.63\%    &  \textbf{71.98$\pm$0.12\%}    &  \textbf{70.66$\pm$0.35\%}    &  69.93$\pm$0.68\%
\\
& FM    &  0$\pm$0\%    &  -3.31$\pm$1.37\%    &  -2.74$\pm$0.49\%    &  -6.50$\pm$1.93\%    &  -16.72$\pm$4.38\%    &  \textbf{0.10$\pm$0.02\%}    &  -1.15$\pm$0.04\%    &  -1.53$\pm$0.33\%    &  -1.44$\pm$0.51\%    &  \textbf{0.49$\pm$0.90\%}    &  -0.31$\pm$0.16\%  
\\
\midrule
\multirow{2}{*}{Amazon Computer}    & PM   &  69.47$\pm$5.83\%    &  66.51$\pm$4.78\%    &  79.35$\pm$1.13\%    &  73.29$\pm$2.33\%    &  70.93$\pm$4.21\%    &  91.62$\pm$1.13\%    &  84.47$\pm$3.91\%    &  90.51$\pm$1.80\%    &  \textbf{95.01$\pm$0.36\%}    &  \textbf{94.72$\pm$1.61\%}    &  93.89$\pm$0.28\%  
\\
& FM     &  0$\pm$0\%    &  -4.85$\pm$2.30\%    &  -3.37$\pm$1.20\%    &  -4.21$\pm$1.98\%    &  -9.72$\pm$4.01\%    &  -1.60$\pm$1.68\%    &  -1.85$\pm$0.17\%    &  -2.74$\pm$1.13\%    &  \textbf{0.92$\pm$0.10\%.}    &  \textbf{0.57$\pm$0.10\%}    &  0.15$\pm$0.10\%  \\

\midrule
\multirow{2}{*}{Amazon Photo}    & PM   &  43.72$\pm$1.30\%    &  81.37$\pm$0.10\%    &  77.72$\pm$2.15\%    &  68.34$\pm$6.96\%    &  40.34$\pm$5.30\%    &  83.66$\pm$0.32\%    &  71.23$\pm$3.04\%    &  86.76$\pm$2.91\%    &  \textbf{94.03$\pm$0.09\%}    &  \textbf{92.55$\pm$0.62\%}    &  91.17$\pm$0.84\%  
\\
& FM     &  0$\pm$0\%    &  -10.86$\pm$3.31\%    &  -4.52$\pm$0.54\%    &  -7.57$\pm$2.31\%    &  -13.98$\pm$3.36\%    &  -5.93$\pm$0.08\%    &  -4.03$\pm$0.23\%    &  -5.93$\pm$0.08\%    &  \textbf{-0.73$\pm$0.18\%}    &  \textbf{-3.66$\pm$0.84\%}    &  -3.79$\pm$0.61\%  \\
\bottomrule
\end{tabular}}
\caption{Performance Mean (PM) and Forgetting Measure (FM) of different methods for node classification. The best two results on each dataset are bolded. We do not consider FM of \textit{Pretrain} because it involves no training on later tasks. OOM means out of memory.}
\label{table:effectiveness}
\end{table*}
\subsubsection{Metrics}
We use performance mean (PM) and forgetting measure (FM) \cite{chaudhry2018riemannian} as metrics. 
\par PM is used to evaluate the learning ability of the model:
\begin{equation}
    PM = \frac{1}{T} \sum_{i=1}^{T} a_{ii},
\end{equation}
where T is the total number of tasks, $a_{ij}$ refers to the model performance on task $j$ after training on task $i$. 
\par FM is used to evaluate the extent of the catastrophic forgetting:
\begin{equation}
    FM = \frac{1}{T-1} \sum_{i=1}^{T-1} a_{Ti} - a_{ii}
\end{equation}
\par In this paper, we use the node classification  accuracy as $a_{ij}$. Therefore, higher PM indicates better overall performance, while more negative FM indicates greater forgetting. In addition, FM$<$0 indicates forgetting, and FM$>$0 indicates no forgetting.


\subsection{Quantitative Results}
\par Table \ref{table:effectiveness} shows the performance of PI-GNN compared with baselines. We make the following observations.

\begin{itemize}[leftmargin=*,noitemsep,topsep=0pt]
    \item In terms of PM, PI-GNN outperforms all continual and dynamic graph learning baselines, indicating that PI-GNN better accommodates new graph patterns by learning stable parameters and splitting new parameters at each snapshot. It should be noted that DNE does not use node attributes and thus has poor performance. 
    \item In terms of FM, dynamic baselines are generally worse than continual baselines as general dynamic graph learning methods fail to address the forgetting problem on previous patterns. Compared to continual graph learning baselines, PI-GNN achieves the least negative FM and thus the least forgetting, indicating PI-GNN's effectiveness in maintaining stable topological patterns.
    \item Although baselines use the same model size as PI-GNN, the informative emerging patterns are learned at the inevitable expense of overwriting the parameters encoding old patterns due to the shared and fixed parameter set. Therefore, their performance is worse than our method \textbf{even when the model sizes are the same}, which validates the effectiveness of explicitly isolating parameters of stable graph patterns and progressively expanding parameters to learn new graph patterns.
    \item Finally, by comparing PI-GNN with Distilled PI-GNN, we observe that Distilled PI-GNN suffers from only $\sim$1\% performance loss and still outperforms baselines, which indicates that the excessive expansion problem is effectively solved by distillation.  
\end{itemize}

\begin{figure}[h]
\centering
\includegraphics[width=0.45\textwidth]{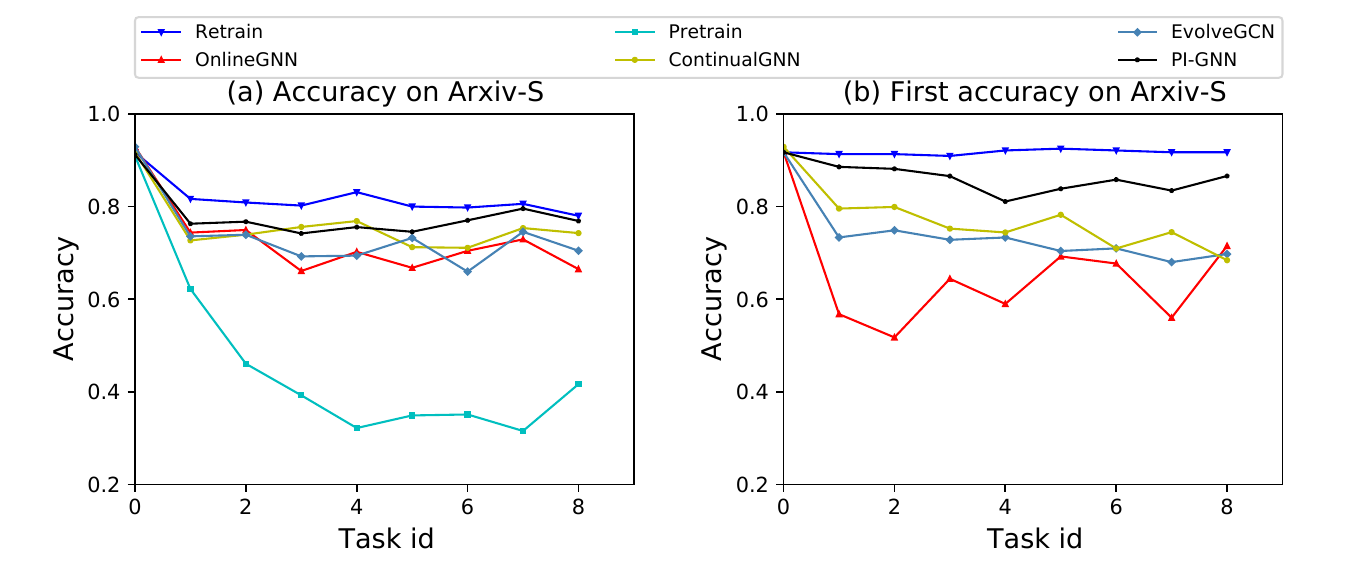}
\caption{Accuracy on current tasks \& task 1 during training.}
\label{figure:effectiveness_acc}
\end{figure}


\begin{table*}[t]
\centering
\begin{tabular}{c|c|cc|cc|cc}
\toprule
                          & Dataset      & \multicolumn{2}{c|}{Arxiv-S}           & \multicolumn{2}{c|}{DBLP-S}        & \multicolumn{2}{c}{Paper100M-S}    \\
\cmidrule{2-8}
                          & Metric             & PM                 & FM                & PM                & FM                & PM               & FM               \\
\midrule
Dynamic                    & EvolveGCN          & 74.08\%.           & -5.55\%           & 61.52\%           & -4.91\%           & 78.14\%          & -3.17\%  \\
\midrule
Continual                  & ContinualGNN       & 75.36\%.           & -2.17\%           & 62.67\%           &  -0.13\%          & 80.59\%          & -2.02\%    \\
\midrule
Upper bound                & Retrain            & 81.86\%.           & -0.12\%           & 64.77\%           & 1.97\%            & 83.42\%          & -0.64\%    \\
\midrule
\multirow{2}{*}{SAGE}     & PI-GNN              & 77.02\%.           & -1.22\%           & 64.75\%           & -0.64\%           & 83.05\%          & -0.93\%    \\
                          & Distilled PI-GNN    & 76.50\%.           & -1.71\%           & 63.49\%           & -1.11\%           & 82.82\%          & -1.49\%    \\
\midrule
\multirow{2}{*}{GCN}      & PI-GNN              & 76.85\%.           & 0.93\%            & 63.62\%           & -0.89\%           & 82.76\%          & -1.35\%    \\
                          & Distilled PI-GNN    & 76.11\%.           & -0.36\%           & 64.09\%           & 0.57\%            & 82.48\%          & -1.92\%    \\
\midrule
\multirow{2}{*}{GAT}      & PI-GNN              & 78.03\%.           & 1.69\%            & 62.78\%           & -1.55\%           & 82.99\%          & -0.75\%    \\
                          & Distilled PI-GNN    & 76.75\%.           & -1.27\%           & 62.81\%           & -1.57\%           & 82.20\%          & -1.11\%    \\
\bottomrule

\end{tabular}
\caption{Performance Mean (PM) and Forgetting Measure (FM) of different methods for node classification.}
\label{table:other_backbone}
\end{table*}

\par We also show the accuracy of PI-GNN and baselines on the current tasks and the first task during the training process in Figure \ref{figure:effectiveness_acc}\footnote{For simplicity, we only show the methods that have best performance among dynamic and continual learning baselines.}. Performance on the current task shows how well the model adapts to new graph patterns, while that on the first task shows how well it maintains knowledge for old graph patterns. We make the following observations:

\begin{figure}[t]
\centering
\includegraphics[width=0.35\textwidth]{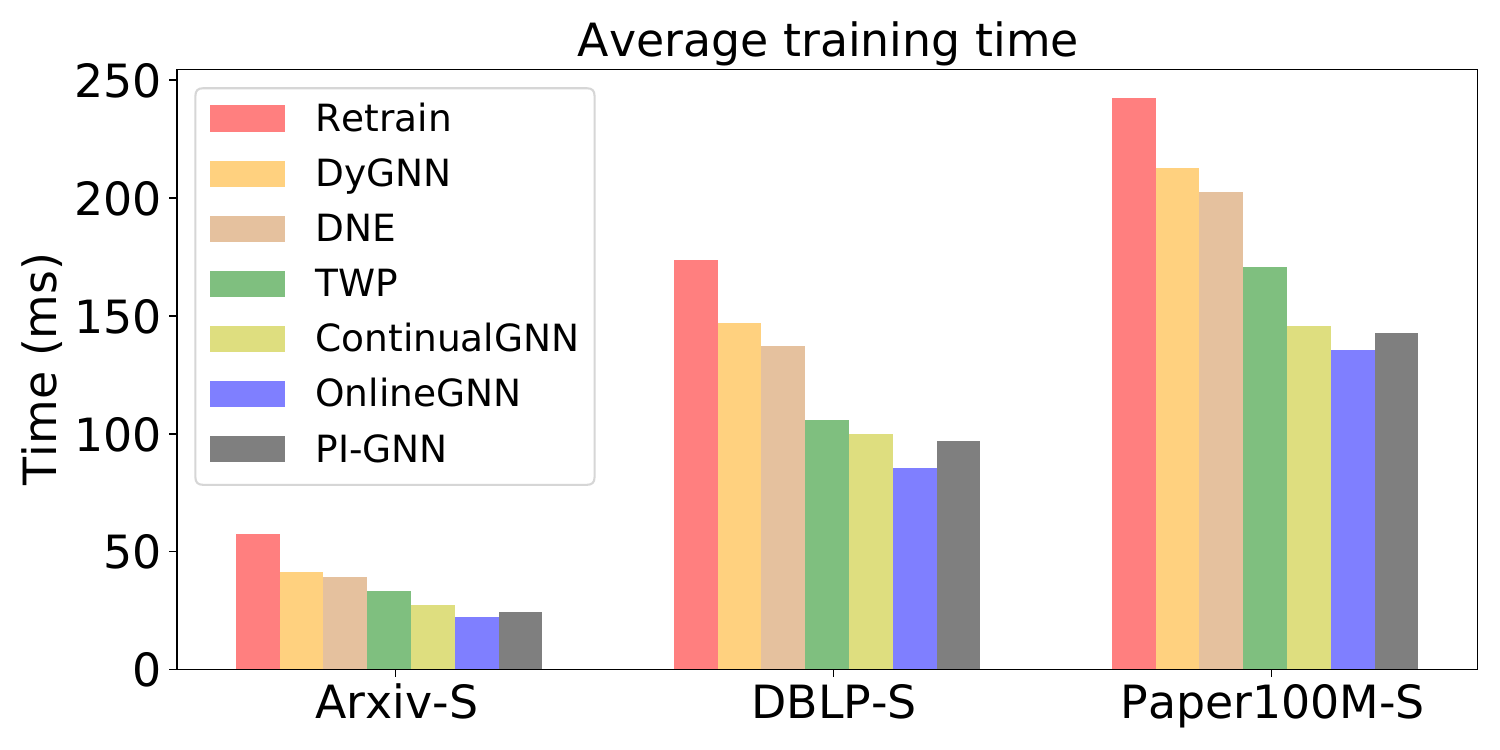}
\caption{Average training time per epoch}
\label{figure:efficiency}
\end{figure}

\begin{itemize}[leftmargin=*,noitemsep,topsep=0pt]
    \item During the training process, PI-GNN is the closest one to \textit{Retrain}, which demonstrates the ability of PI-GNN to adapt to new patterns and coincides with our theoretical analysis. \textit{OnlineGNN} fluctuates greatly because it rewrites parameters and thus suffers from the catastrophic forgetting problem. \textit{Pretrain} cannot learn new patterns thus it performs worst. 
    \item The performance of \textit{ContinualGNN} and \textit{EvolveGCN} gradually drops. We attribute this to the following reasons. \textit{ContinualGNN} gradually forgets the old knowledge after learning many new patterns because of the constant model capacity, which makes it difficult to accommodate the new patterns using existing parameters without forgetting. \textit{EvolveGCN} can learn on the whole snapshot, but it aims at capturing dynamism of the dynamic graph and will forget inevitably.
\end{itemize}

\subsection{Efficiency}
\par Figure \ref{figure:efficiency} presents the average training time of PI-GNN per epoch compared with baselines. We do not show the average training time of \textit{EvolveGCN} because it uses the whole snapshots and RNN to evolve the parameters, which needs more overhead than \textit{Retrain}. \textit{DiCGRL} is not shown as it suffers from out of memory problem in Paper100M-S. From the results, we observe that \textit{TWP} has to calculate the topology weights to preserve structure information and thus results in more overheads. It worth noting that \textit{DyGNN} and \textit{DNE} have different settings where they update the nodes embedding after every new edge comes, therefore they have relative large overheads. PI-GNN is as efficient as \textit{OnlineGNN}. The high efficiency can be attributed to sampling of $G^{t-1}_{memory}$ and incremental learning on $\Delta G^t$ and $G_{memory}^{t-1}$ at each snapshot. 

\begin{figure}[t]
\centering
\subfigure[PM for different rectification epochs.]{
\includegraphics[width=0.18\textwidth]{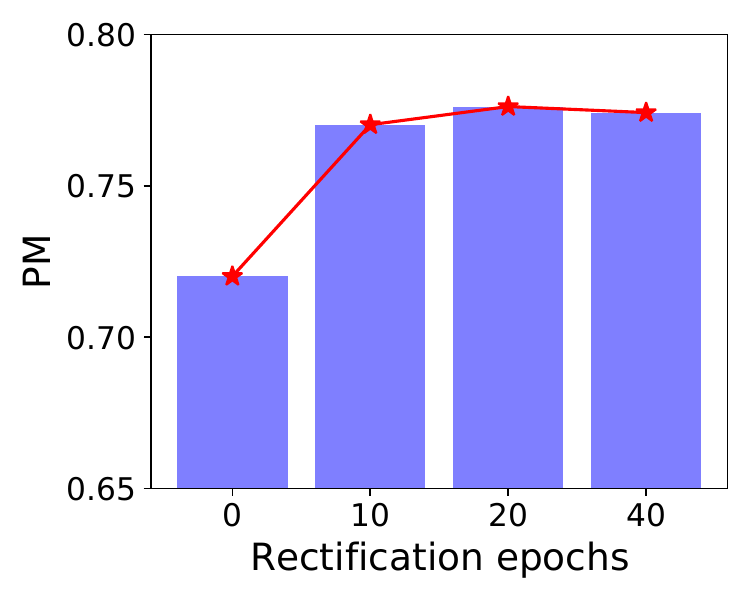}
\label{figure:rectification}
}
\subfigure[Ablation Study on Isolation and Expansion]{
\includegraphics[width=0.18\textwidth]{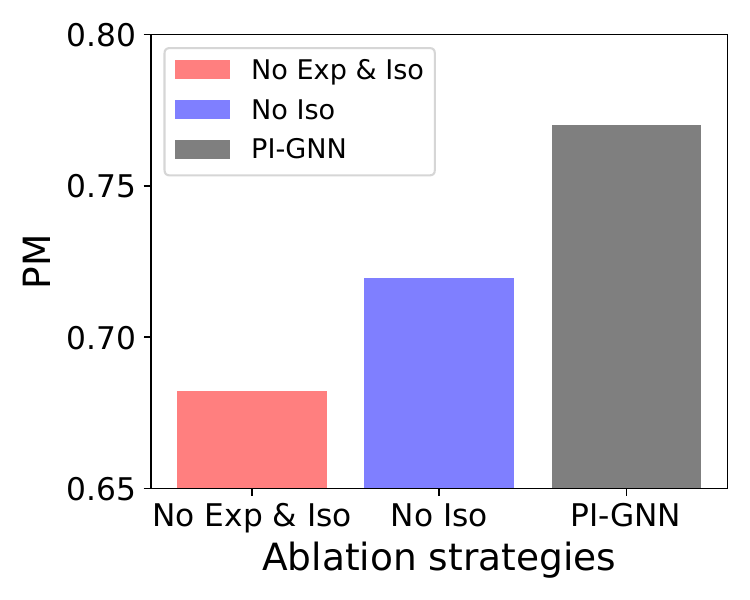}
\label{figure:isolation}
}

\subfigure[Sample Size of $G_{memory}^{t-1}$]{
\includegraphics[width=0.20\textwidth]{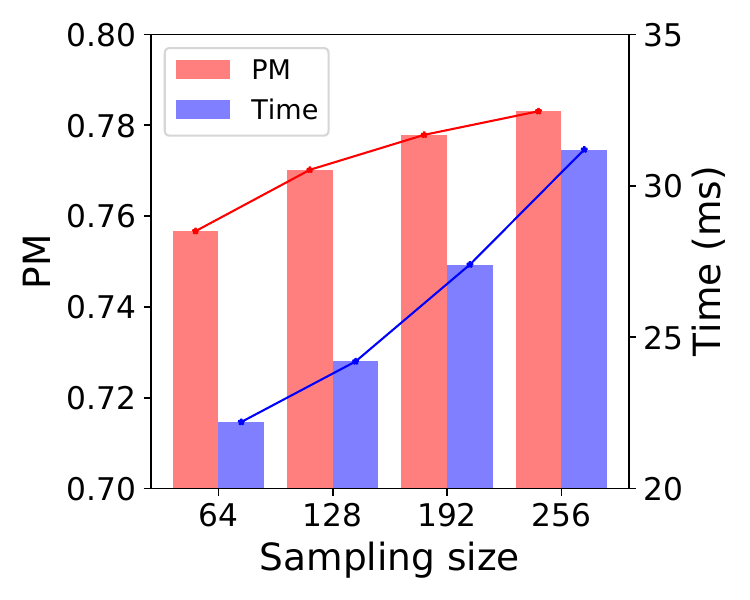}
\label{figure:sample_num}
}
\subfigure[Expansion Units]{
\includegraphics[width=0.20\textwidth]{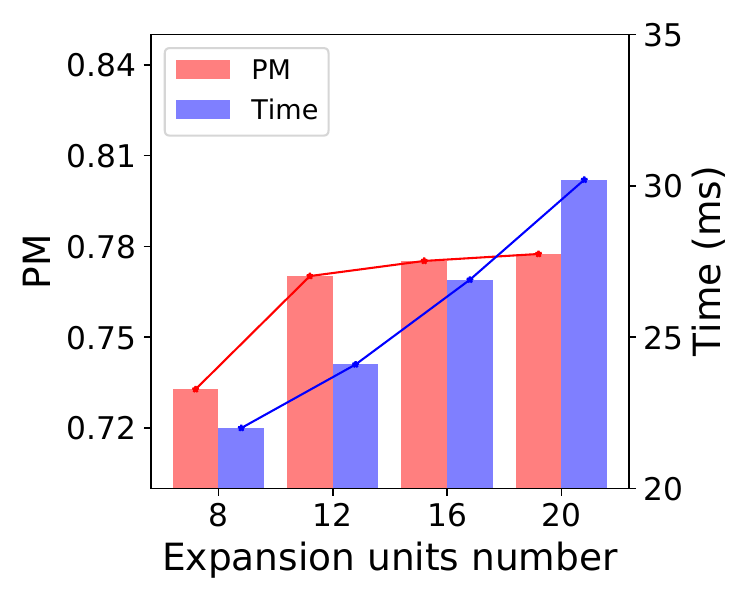}
\label{figure:expansion_num}
}
\caption{Analysis Results of PI-GNN.}
\label{figure:model_study}
\end{figure}

\subsection{Model Analysis}
\subsubsection{Impact of different backbones}
\par We use different GNN models as the backbones of PI-GNN. The results can be found in Table \ref{table:other_backbone}. We only show \textit{ContinualGNN} and \textit{EvolveGCN} which have best performance in continual learning methods and dynamic learning methods respectively. We find that our PI-GNN can be combined with different GNN backbones properly.

\subsubsection{Knowledge Rectification and Parameter Isolation}
First, we study the impact of knowledge rectification. We train different epochs for the knowledge rectification stage and show the results in Figure \ref{figure:rectification}. On one hand, we observe a large degradation upon the absence of knowledge rectification (0 epochs), as we may freeze parameters for unstable patterns. On the other hand, the number of epochs has little effect as long as rectification is present. 

Second, we perform an ablation study on the parameter isolation stage and show results in Figure \ref{figure:isolation}. The significant performance drop upon removal of either parameter isolation or expansion underscores their contribution to PI-GNN.




\subsubsection{Hyperparameter Study}
We conduct detailed hyperparameter studies on the hyperparameter of our model.

\noindent\textbf{Sampling Size Study.} We first study the size of sampling $G_{memory}^{t-1}$. The results are shown in Figure \ref{figure:sample_num}. We find that sampling more nodes can ensure the tightness of the optimization upper bound. When the upper bound is tight enough, sampling abundant nodes no longer brings additional benefits, but increases the overhead. Therefore, we need to choose a suitable balance between sampling and efficiency. Empirically, we find that good results can be achieved when sampling ~1\% of the total number of nodes. In our experiments, we sampling 128 nodes for Arxiv-S and 256 for the other datasets empirically (about 1\%).

\noindent\textbf{Expansion Units Number Study.}
We study the influence of expansion units number. As shown in Figure \ref{figure:expansion_num}. We find that a small number (e.g. 12) of units is sufficient as excessive expansion leads to little performance improvement but greater overhead.

\noindent\textbf{Balance Factor Study.} We show the influence of balance factor $\beta$ in Table \ref{table:appendix_beta_influence} and $\alpha$ in Table \ref{table:appendix_lambda_influence}. We find that improper balance will have some negative influence on the performance.



\begin{table}[!t]
\begin{minipage}[!t]{0.48\columnwidth}
  \renewcommand{\arraystretch}{1.3}
  \centering
  
  \begin{tabular}{c|c|c}
\toprule
$\beta$   & PM        & FM \\
\midrule
0               & 75.96\%   & -1.17\%   \\
0.01            & 77.02\%   & -0.22\%   \\
0.02            & 76.95\%   & -1.61\%   \\
0.04            & 76.99\%   & -0.87\%   \\
\midrule
\end{tabular}
\caption{Influence of $\beta$.}
  \label{table:appendix_beta_influence}
  \end{minipage}
\begin{minipage}[!t]{0.48\columnwidth}
  \renewcommand{\arraystretch}{1.3}
  
  \centering
  
  \begin{tabular}{c|c|c}
\toprule
$\lambda$   & PM        & FM \\
\midrule
0              & 75.24\%   & 1.46\%    \\
0.1            & 77.02\%   & -1.22\%   \\
0.2            & 77.22\%   & -0.59\%   \\
0.4            & 76.32\%   & -1.25\%   \\
\midrule
\end{tabular}
\caption{Influence of $\lambda$.}
  \label{table:appendix_lambda_influence}
  \end{minipage}
\end{table}

\subsection{Layer Output Visualization}
\begin{figure}[t]
\centering
\includegraphics[width=0.35\textwidth]{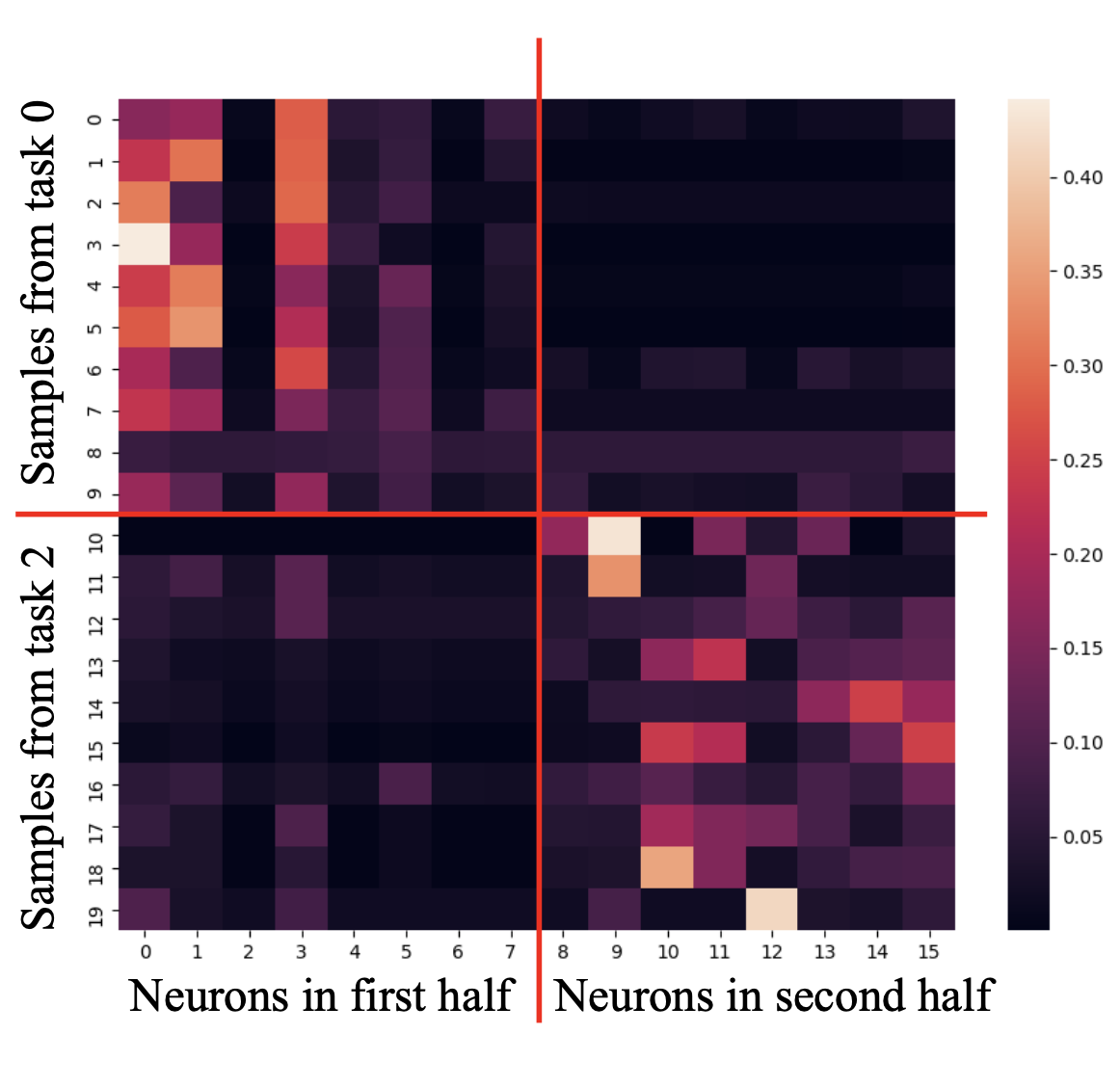}
\caption{First layer output of samples from different tasks}
\label{figure:ablation_stage2}
\end{figure}

To verify our argument that different parameters contribute to capturing different patterns, which serves as the foundation of PI-GNN, we visualize the first layer output of PI-GNN on Arxiv-S dataset. Specifically, we use 20 samples from task 0 and task 2 respectively to show the relationship between different patterns and the first layer output of our model. As shown in Figure~\ref{figure:ablation_stage2}, the activated neurons are mainly concentrated in the first half for the samples from task 0, and the activated neurons are mainly concentrated in the second half for the samples from task 2, indicating that different parameters contribute to different patterns.

\begin{table}[t]
\centering
\small
\begin{tabular}{c|c|c|c}
\toprule
\multicolumn{1}{l|}{} & Arxiv-S   & DBLP-S   & Paper100M-S \\
\midrule
PI-GNN Size                & 213KB     & 759KB.    & 285KB \\
Distilled PI-GNN Size      & 47KB      & 47KB.     & 47KB \\
\bottomrule
\end{tabular}
\caption{Model size before/after distillation}
\label{table:appendix_distillation}
\end{table}

\subsection{Model Compression}
Finally, we compare PI-GNN before and after compression along with some baselines. The results are reported in Table~\ref{table:appendix_distillation}. We observe that after compression, our model size is reduced by 4× with less than 1\% performance drop (shown in Table~\ref{table:effectiveness}). The results show that PI-GNN can be compressed to avoid being excessively expanded, and thus, PI-GNN is efficient even under large $T$.

\section{Related Work}
\subsection{Dynamic Graph Learning}
\par Most real-world graphs are dynamic with new nodes and edges. Many efforts have been paid to learn representations for dynamic graphs~\cite{DBLP:journals/corr/abs-1805-11273,ijcai2019-640,DBLP:conf/nips/HajiramezanaliH19,DBLP:conf/kdd/KumarZL19}. DNE~\cite{du2018dynamic} and DyGNN~\cite{ma2020streaming} incrementally update node embeddings to reflect changes in graphs. Dyrep~\cite{trivedi2019dyrep} models the micro and macro processes of graph dynamics and fuse temporal information to update node embeddings. EvolveGCN~\cite{pareja2020evolvegcn} uses recurrent networks to model evolving GNN parameters corresponding to the graph dynamics. However, dynamic graph learning methods in general focus on learning better embeddings for the \textit{current} and \textit{future} graphs, rather than \textit{previous} ones. Consequently, they commonly fail to maintain performance on previous graphs, which is known as \textit{catastrophic forgetting}. 

\subsection{Continual Graph Learning}

\par To alleviate catastrophic forgetting in dynamic graphs, researchers propose \textit{continual graph learning} methods. ER-GNN~\cite{zhou2021overcoming} generalizes replay-based methods to GNNs. ContinualGNN~\cite{wang2020streaming} proposes importance sampling to store old data for replay and regularization. DiCGRL~\cite{kou2020disentangle} disentangles node representations and only updates the most relevant parts when learning new patterns. TWP~\cite{liu2021overcoming} considers graph structures and preserves them to prevent forgetting. However, all these methods aim to maintain performance on previous graphs while learning for future ones with the same set of parameters with a fixed size. Consequently, a tradeoff between old and new patterns in dynamic graphs has to be made, rendering existing methods sub-optimal.



\section{Conclusion}
\par In this paper, we propose PI-GNN for continual learning on dynamic graphs. The idea of PI-GNN naturally fits the unique challenge of continual graph learning. We specifically find parameters that learn stable graph patterns via optimization and freeze them to prevent catastrophic forgetting. Then we expand model parameters to incrementally learn new graph patterns. Our PI-GNN can be combined with different GNN backbones properly. We theoretically prove that PI-GNN optimizes a tight upper bound of the retraining loss on the entire graph. PI-GNN is tested on eight real-world datasets and the results show its superiority.


\clearpage
\appendix
\section{Appendix}
\subsection{Proof of Lemma \ref{lem}}
\label{app:lemma_proof}
We show the proof of Lemma \ref{lem} in this subsection. We first re-state Lemma \ref{lem}:
\par \textit{$L(f_{\theta_{1}}(G)+f_{\theta_{2}}(G)) \leq \frac{1}{2}L(f_{\theta_{1}}(G)) + \frac{1}{2}L(f_{\theta_{2}}(G))$ where the equality can be established if $f_{\theta_{1}}$ and $f_{\theta_{2}}$ have the same output on $G$.}
\begin{proof}
This lemma largely follows from the convexity of $L$ (i.e. cross entropy) and Jensen’s inequality. Here we prove it in detail. For simplicity, we assume that the output dimension of $f_{\theta_{1}}$ and $f_{\theta_{2}}$ are both 2, which can be easily generalized to higher dimensionality. We denote the output of $f_{\theta_{1}}(G)$ as $H_{1}$ where $h^{i}_{1} = (h^{i}_{1,1}, h^{i}_{1,2})$ and $h^{i}_{1}$ is the i-th row of $H_{1}$. Similarly, we denote the output of $f_{\theta_{2}}(G)$ as $H_{2}$ where $h^{i}_{2} = (h^{i}_{2,1}, h^{i}_{2,2})$and $h^{i}_{2}$ is the i-th row of $H_{2}$
\par The definition of $L(f_{\theta}(G))$ can be derived as:
\begin{equation}
\begin{aligned}
 L \left (f_{\theta}\left(G\right) \right )
    = & \sum_{h^{i} \in H} \mathrm{CrossEnt}\left (\mathrm{softmax} \left (h^{i} \right ), y^{i} \right ) \\
    = & \sum_{h^{i} \in H} \left\{ -y^{i}_{1}\log\left(\frac{e^{h^{i}_{1}}}{e^{h^{i}_{1}} + e^{h^{i}_{2}}}\right) \right.  \left. -y^{i}_{2}\log\left(\frac{e^{h^{i}_{2}}}{e^{h^{i}_{1}} + e^{h^{i}_{2}}}\right) \right\}
\label{raw_definition}
\end{aligned}
\end{equation}
\par Therefore if we dilate the inequality of Lemma \ref{lem} according to Equation \ref{raw_definition} we will get the formulation which we need to prove as follow:
\begin{equation}
\begin{aligned}
    & 2 \sum_{h_{1}^{i} \in H_{1}, h_{2}^{i} \in H_{2}} \left\{ -y^{i}_{1}\log\left(\frac{e^{h^{i}_{1,1} + h^{i}_{2,1}}}{e^{h^{i}_{1,1} + h^{i}_{2,1}} + e^{h^{i}_{1,2} + h^{i}_{2,2}}}\right) \right. \\ & \left. -y^{i}_{2}\log\left(\frac{e^{h^{i}_{1,2} + h^{i}_{2,2}}}{e^{h^{i}_{1,1} + h^{i}_{2,1}} + e^{h^{i}_{1,2} + h^{i}_{2,2}}}\right)\right\} \\
    \leq \\
    & \sum_{h_{1}^{i} \in H_{1}} \left\{ -y^{i}_{1}\log\left(\frac{e^{h^{i}_{1,1}}}{e^{h^{i}_{1,1}} + e^{h^{i}_{1,2}}}\right) \right. \left. -y^{i}_{2}\log\left(\frac{e^{h^{i}_{1,2}}}{e^{h^{i}_{1,1}} + e^{h^{i}_{1,2}}}\right) \right\} \\
    & + \sum_{h_{2}^{i} \in H_{2}} \left\{ -y^{i}_{1}\log\left(\frac{e^{h^{i}_{2,1}}}{e^{h^{i}_{2,1}} + e^{h^{i}_{2,2}}}\right) \right.  \left. -y^{i}_{2}\log\left(\frac{e^{h^{i}_{2,2}}}{e^{h^{i}_{2,1}} + e^{h^{i}_{2,2}}}\right) \right\}
\end{aligned}
\end{equation}
\par Next we prove that for node $i$ this relation holds where we omit $i$ in the following notation, for simplicity, We denote the $h_{1,1}, h_{1,2}, h_{2,1}, h_{2,2}$ as $h_{11}, h_{12}, h_{21}, h_{22}$ respectively. Thus we need prove:
\begin{equation}
\begin{aligned}
    & -2y_{1}\log\left(\frac{e^{h_{11} + h_{21}}}{e^{h_{11} + h_{21}} + e^{h_{12} + h_{22}}}\right)  -2y_{2}\log\left(\frac{e^{h_{12} + h_{22}}}{e^{h_{11} + h_{21}} + e^{h_{12} + h_{22}}}\right) \\
    \leq & -y_{1}\log\left(\frac{e^{h_{11}}}{e^{h_{11}} + e^{h_{12}}}\right) -y_{2}\log\left(\frac{e^{h_{12}}}{e^{h_{11}} + e^{h_{12}}}\right) \\
    & -y_{1}\log\left(\frac{e^{h_{21}}}{e^{h_{21}} + e^{h_{22}}}\right) -y_{2}\log\left(\frac{e^{h_{22}}}{e^{h_{21}} + e^{h_{22}}}\right) \\
    \Leftrightarrow \\
\end{aligned}
\end{equation}
\begin{equation}
\begin{aligned}
    & y_{1}\log\left(\frac{\left(e^{h_{11} + h_{21}}\right)^{2}}{\left(e^{h_{11} + h_{21}} + e^{h_{12} + h_{22}}\right)^{2}}\right)  + y_{2}\log\left(\frac{\left(e^{h_{12} + h_{22}}\right)^{2}}{\left(e^{h_{11} + h_{21}} + e^{h_{12} + h_{22}}\right)^{2}}\right) \\
    & - y_{1}\log\left(\frac{e^{h_{11}+h_{21}}}{e^{h_{11}+h_{21}} + e^{h_{12}+h_{21}} + e^{h_{11}+h_{22}} + e^{h_{12}+h_{22}}}\right) \\ & -y_{2}\log\left(\frac{e^{h_{12}+h_{22}}}{e^{h_{11}+h_{21}} + e^{h_{12}+h_{21}} + e^{h_{11}+h_{22}} + e^{h_{12}+h_{22}}}\right)  \geq 0 \\
    \Leftrightarrow \\
    & y_{1}\log\left(\frac{1 + \frac{e^{h_{12}+h_{21}}+e^{h_{11}+h_{22}}}{e^{h_{11}+h_{21}}+e^{h_{12}+h_{22}}}}{1+\frac{e^{h_{12}+h_{22}}}{e^{h_{11}+h_{21}}}}\right) + y_{2}\log\left(\frac{1 + \frac{e^{h_{12}+h_{21}}+e^{h_{11}+h_{22}}}{e^{h_{11}+h_{21}}+e^{h_{12}+h_{22}}}}{1+\frac{e^{h_{11}+h_{21}}}{e^{h_{12}+h_{22}}}}\right) \geq 0
\end{aligned}
\end{equation}

\par We assume that $\frac{e^{h_{12}}}{e^{h_{11}}} = m$, $\frac{e^{h_{22}}}{e^{h_{21}}} = n$, we should prove that:
\begin{equation}
\begin{aligned}
    y_{1}\log\left(\frac{1+\frac{m/n+n/m}{1+mn}}{1+mn}\right) + y_{2}\log\left(\frac{1+\frac{m/n+n/m}{1+mn}}{1+\frac{1}{mn}}\right) \geq 0
\end{aligned}
\end{equation}

\par We note that node $i$ can be either in class $y_{1}$ or in class $y_{2}$, Thus:
\par (1) If node $i$ is labeled with $y_{1}$, which means $y_{1}=1$ and $y_{2}=0$, we have to prove that:
\begin{equation}
    \log\left(\frac{1+\frac{m/n+n/m}{1+mn}}{1+mn}\right) \geq 0
    \label{condition_1}
\end{equation}
\par After optimization we can classify node $i$ rightly, which means that $m \leq 1$ and $n \leq 1$, and thus:
\begin{equation}
\begin{aligned}
    \frac{m/n+n/m}{1+mn} & \geq 1 \\
    mn & \leq 1
\end{aligned}
\end{equation}
\par Therefore, inequality \ref{condition_1} is proved.

\par (2) If node $i$ is labeled with $y_{2}$, which means $y_{1}=0$ and $y_{2}=1$, we have to prove that:
\begin{equation}
    \log\left(\frac{1+\frac{m/n+n/m}{1+mn}}{1+\frac{1}{mn}}\right) \geq 0
    \label{condition_2}
\end{equation}
\par After optimization we can classify node $i$ rightly, which means that $m \geq 1$ and $n \geq 1$, and thus:
\begin{equation}
\begin{aligned}
    & m^2 + n^2 \geq 1 + mn \\
    \Leftrightarrow & \frac{n}{m} + \frac{m}{n} \geq 1 + \frac{1}{mn} \\
    \Leftrightarrow & \frac{m/n+n/m}{1+mn} \geq \frac{1}{mn} \\
\end{aligned}
\end{equation}
\par Therefore, inequality \ref{condition_2} is proved.
\par In addition, if $m=n$, which means $f_{\theta_{1}}(G)$ and $f_{\theta_{1}}(G)$ have the same output on node $i$, we have:
\begin{equation}
\begin{aligned}
    y_{1}\log\left(\frac{1+\frac{m/n+n/m}{1+mn}}{1+mn}\right) + y_{2}\log\left(\frac{1+\frac{m/n+n/m}{1+mn}}{1+\frac{1}{mn}}\right) = 0
\end{aligned}
\end{equation}
\par Therefore, Lemma \ref{lem} is proved.
\end{proof}

\bibliographystyle{ACM-Reference-Format}
\balance
\bibliography{sample-base}

\clearpage

\end{document}